\documentclass{bmvc2k}

\usepackage{kotex}
\usepackage{multirow}

\usepackage{url}
\usepackage{amsfonts}

\usepackage[normalem]{ulem} 

\newif\ifdraft\draftfalse
\ifdraft

\else

\newcommand{\etal}{\textit{et al}. }

\usepackage{algorithm}
\usepackage{algpseudocode}
\usepackage{mathtools,cases,xparse,eqparbox}
\usepackage{bbold}

\fi

\title{Font Representation Learning via Paired-glyph Matching}

\addauthor{Junho Cho }{junhocho@snu.ac.kr}{12}
\addauthor{Kyuewang Lee}{kyuewang@snu.ac.kr}{1}
\addauthor{Jin Young Choi}{jychoi@snu.ac.kr}{1}

\addinstitution{
 Department of ECE, ASRI,\\
 Seoul National University,\\
 Seoul, Korea
}
\addinstitution{
 Samsung Advanced Institute of Technology, Samsung Electronics, \\
 Suwon, Korea
}

\runninghead{Cho et al.}{Paired-glyph matching learning}

\def\etal{\emph{et al}\bmvaOneDot}

\definecolor{orange}{rgb}{1,0.5,0}
\definecolor{blue}{rgb}{0,0,0.6}
\definecolor{green}{rgb}{0,0.6,0}

\usepackage{xcolor}
\begin{document}

\maketitle

\begin{abstract}
Fonts can convey profound meanings of words in various forms of glyphs.
Without typography knowledge, manually selecting an appropriate font or designing a new font is a tedious and painful task.
To allow users to explore vast font styles and create new font styles, font retrieval and font style transfer methods have been proposed.
These tasks increase the need for learning high-quality font representations.
Therefore, we propose a novel font representation learning scheme to embed font styles into the latent space. 
For the discriminative representation of a font from others, we propose a paired-glyph matching-based font representation learning  model that attracts the representations of glyphs in the same font to one another, but pushes away those of other fonts.
Through evaluations on font retrieval with query glyphs on new fonts, we show our font representation learning scheme achieves better generalization performance than the existing font representation learning techniques.
Finally on the downstream font style transfer and generation tasks, we confirm the benefits of transfer learning with the proposed method. 

\end{abstract}

\section{Introduction}

A font, which is a graphical representation of text, delivers certain visual feelings in multimedia through its matching style set of glyphs.
Professional designers carefully choose fonts to convey their design intent.
However, it is challenging to search for a specific font in the vast number of fonts available.
Moreover, designing fonts requires typography knowledge, and aspiring designers can take months to learn typography.
To cope with these difficulties, fonts should be easier to search for and create.
There has been active research on font retrieval~\cite{odonovan:2014:font, Chen2019LargeScaleTF, font-retrieval-icip2010, fonts-like-this-happier-acmmm2020, kang2021shared}, font style transfer and generation~\cite{DG-Font, zhang2018separating, mcgan, Hayashi2019GlyphGANSF}.

\begin{figure}[t!]
  \centering
  \includegraphics[width=\linewidth]{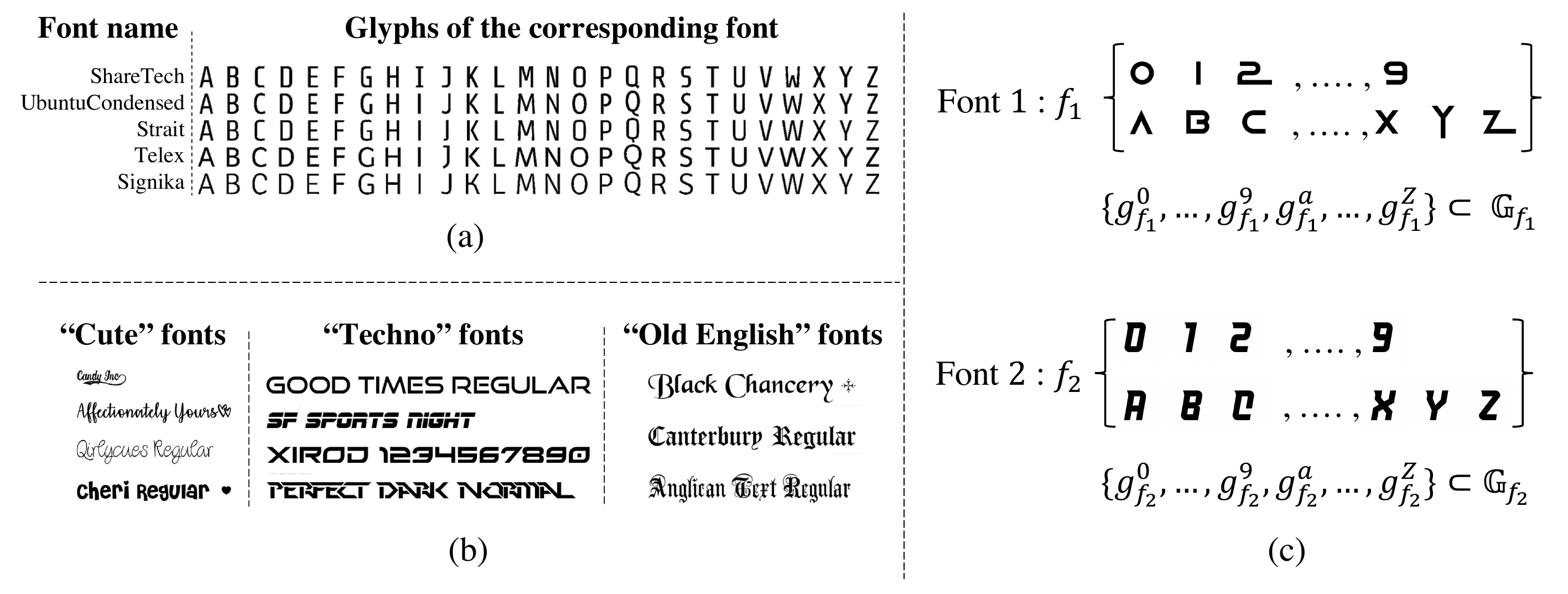}
  \vspace{-0.9cm}
  \caption{
  (a) Challenging fonts to distinguish. (b) Various fonts tagged as "Cute", "Techno" and "Old English", (c) Notations of glyphs $g$ and glyph set  $\mathbb{G}_{f}$ for font $f$}
\vspace{-0.5cm}
  \label{fig:teaser}
\end{figure}

Font retrieval is a task that allows users to find similar looking fonts.
Users can browse the fonts in the latent space to find the font they want.
Through recognizing font style and generating new glyphs with the corresponding style, font style transfer and generation can ease the labor-intensive job of creating numerous glyphs with a certain font style.
Font retrieval, style transfer and generation have historically focused on their own specific goals.
However, if a powerful font representation learning method is devised, these tasks are considered downstream tasks, and performance gains can be expected through transfer learning~\cite{marcelino2018transfer}.
Therefore, we present a novel font representation learning scheme for the broader generalization on font-related downstream tasks.
However, learning fonts is not as easy as one might think.
Five fonts shown in Figure~\ref{fig:teaser} (a), ShareTech, UbuntuCondensed, Strait, Telex, Signika are very difficult to distinguish with our eyes.
Unlike general objects with textures~\cite{geirhos2018imagenettrained}, fonts have typographic elements~(e.g., cap, x-height, serif, stem, stroke, descender, ascender, aperture) which are shape-based representations.
Therefore, distinguishing these nuances is important for learning high-quality font representations.

In this paper, to mitigate the aforementioned difficulties, we approach how to learn these nuances through pairwise glyph similarity learning. 
More specifically, we try to learn the style representation of a font regardless of the shape of the character.
That is, each font style keeps its unique nuance though the glyphs in the font have diverse shapes, which is referred to as {\it Glyph-font-consistency}.
Paying attention to this unique nuance, we propose a new representation learning scheme to learn font features, keeping {\it Glyph-font-consistency} through a paired-glyph matching strategy.
The proposed scheme attracts the font representations of glyphs in the same font to one another, but pushes away those of other fonts.
We study generalization ability of our discriminative font representation learning scheme compared to existing font representation learning techniques.
Finally, we evaluate  performance improvement by transfer learning of our font representation learning scheme in the downstream font style transfer and generation tasks.

\section{Related Works}

\subsection{Font Classification \& Retrieval}

Font classification and recognition models~\cite{fontcls1, fontcls2, fontcls3, fontcls4,  font-cls} are used to increase performance in text detection and recognition~\cite{boost-text-recog, boost-text-recog2}, to make difficult calligraphy easier for users to recognize~\cite{fontcls5}.
These methods of font classification only work with fixed sets of fonts, so they lack generalization to countless number of unseen fonts.
Therefore, various retrieval-based methods~\cite{odonovan:2014:font, Chen2019LargeScaleTF, font-retrieval-icip2010, fonts-like-this-happier-acmmm2020, kang2021shared} have been proposed  for learning font representation and various related applications.
Before the deep learning-based method appeared, Kataria~\etal~\cite{font-retrieval-icip2010} extracted the SIFT~(Scale-Invariant Feature Transform)~\cite{SIFT} feature from each glyph of the font and defined the concatenation of glyphs as the font embedding. 
O'Donovan~\etal~\cite{odonovan:2014:font} defined the attributes (e.g., artistic, attractive, pretentious) of fonts and used crowd-sourced way to annotate the font attributes.
And by learning a model to predict the attributes of fonts, O'Donovan~\etal predicted attributes even for unseen fonts.
However, specifying font attributes and determining their values is a rather subjective task, and the cost of annotations is very high, which limited  annotations for small number of  fonts.
In light of this, tag-based font retrieval websites with relatively low annotation costs (e.g., \url{dafont.com}, \url{myfonts.com}, \url{10001fonts.com}) appeared.

These websites provide a tag-based font search service that allows users to select and download selected fonts.
Figure~\ref{fig:teaser} (b) shows how users can search for fonts  based on a query (e.g., cute, techno, Old English).
However, the tag-based font search has the disadvantage that, much like the problem with tag-based image searches, the tag does not sufficiently describe the font, and even appropriate tags may be subjective. 
With the advent of deep learning, some tag-based font retrieval studies~\cite{Chen2019LargeScaleTF, kang2021shared, fonts-like-this-happier-acmmm2020} have tried to associate font tags to learn font representation in a data-driven manner.
These studies proposed a method to perform tag classification~\cite{Chen2019LargeScaleTF} on fonts or to share the font latent space with the tag representation through Word2vec~\cite{word2vec, kang2021shared, fonts-like-this-happier-acmmm2020}.
They investigated how the specific glyph shape of a font was related to a specific emotional font tag.
However, these methods cannot learn font embedding without font tags.


\subsection{Font Style Transfer \& Font Generation}

The necessity of font style transfer methods comes from the tedious and labor-intensive job of creating numerous glyphs with font style.
For example, Chinese contains more than 60,000 characters and Korean contains 11,172 characters.
Early font style transfer methods~\cite{zi2zi, Xi2020JointFontGANJG, Hassan2021UnpairedFF, DG-Font} were based on image-to-image translation models~\cite{pix2pix2017, acgan, dtn}  with the advance of generative adversarial networks~\cite{gan_ian}.
These methods transferred the font style of one glyph image to another glyph image.
These methods typically extracted font style features from glyph images for reference via a font style encoder model.
Each method focused on the structural design of the font style encoder, because the style encoder needed to learn a good font representation so  the font style was represented well in the output image.
That is, better font representation learning was helpful for better quality font generation.

\section{Methodology}

\subsection{Notations and Our Research Objective}

To establish appropriate context, it is important to outline how we denote characters, glyphs and fonts.
A character set is defined by a class of characters, for instance, $\mathbb{C}_\text{0-9}  = \{  0, 1, 2, ..., 9 \}$,  $\mathbb{C}_\text{a-Z} = \{  a, b, c, ... , X ,Y, Z  \}$ and $\mathbb{C}_\text{0-Z} = \{  0, 1, 2, ..., 9, a, b, c, ... , X ,Y, Z  \}$.
A glyph is an image form of a character that has a specific style in a font.
For example, if a glyph describes the character ``Z'' with a certain font $f_1$,  we denote the glyph as $g_{f_1}^Z$.
Figure~\ref{fig:teaser}~(c) shows that a font $f_1$ includes a matched set of glyphs for a character set $\mathbb{C}$.
For example, the glyph set with font $f_1$ of character set $\mathbb{C}_\text{0-Z} $ 
is denoted by
\begin{align}
\mathbb{G}_{f_1}^{\mathbb{C}_\text{0-Z}}=\{ g_{f_1}^c | c \in \mathbb{C}_\text{0-Z} \} =
\{   g_{f_1}^0, g_{f_1}^1, g_{f_1}^2, ..., g_{f_1}^X, g_{f_1}^Y, g_{f_1}^Z\} \subset \mathbb{G}_{f_1}.
\end{align}
Denoting the set of all fonts in the world by $\mathbb{F}$, two different fonts $f_i, f_{j \ne i} \in \mathbb{F}$ convey different styles through two glyph sets ($\mathbb{G}_{f_i}=\{g_{f_i}^{c} | c \in \mathbb{C} \}$ and $\mathbb{G}_{f_j}=\{g_{f_j} ^{c} | c \in \mathbb{C}\}$).

\begin{figure}[t!]
  \centering
  \includegraphics[width=1\linewidth]{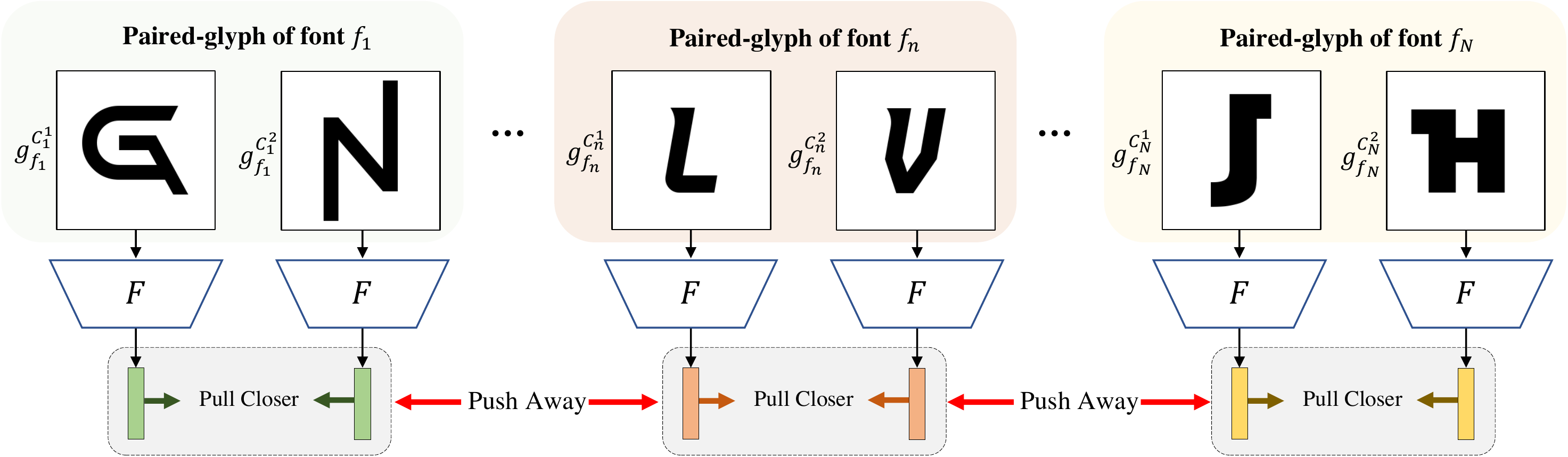}
  \caption{
  Overall scheme of Paired-glyph Matching learning
  }
  \label{fig:framework}
  \vspace{-0.5cm}
\end{figure}

Based on the intrinsic relationship between fonts and glyphs, our research objective is to embed the fonts to representation space so that the  glyphs in the same font are embedded into a small representation area far from those of the other fonts.
To this end, we propose a {\it Paired-glyph Matching} learning scheme  to pull the font representations of all glyphs in $\mathbb{G}_{f_i}$  closer to one another but push away from the font representations of  
the glyphs in the other glyph sets $\mathbb{G}_{f_{j \ne i}}$ and vice versa, as shown in Figure~\ref{fig:framework}.

\subsection{Paired-glyph Matching Learning}

In {\it Paired-glyph Matching} learning, 
we randomly sample two fonts, $f_1$ and $f_2$, and two characters, $c_1$ and $c_2$.
Then, we get a set of four glyphs $\{g_{f_t}^{c_i}| t=1,2; i=1,2\}$ expressing the font $f_t$ for the character $c_i$.
For the objective function to train $F$, we use cosine similarity given by $\texttt{sim}(u,v) = \frac{u^{T}v}{\|u\|\|v\|}$ as the dot product between L2 normalized $u$ and $v$, where $u,v$ are the font representations.
We train the model $F$ to map the glyphs from the same font into similar representations and  those from different fonts into discriminative representations. That is,
we maximize $\texttt{sim}(F(g_{f_t}^{c_1}), F(g_{f_t}^{c_2})), t=1,2$ and minimize $\texttt{sim} (F(g_{f_1}^{c_i}), F(g_{f_2}^{c_i})), i=1,2$.
Glyphs of the same character look alike in the image space, even though their fonts are different from one another.
However, the aforementioned objective drives the different font glyphs of the same character to be embedded far away from one another in the latent space.
That is, we train the model $F$ to focus on the font style of a glyph more than the shape of a character.

To generalize {\it Paired-glyph Matching} with a minibatch of $N$ fonts, we randomly sample fonts $\{f_1, f_2, ..., f_{N}\}$ from the training set.
We randomly sample two different glyph images for each font as 
$\{\{g_{f_n}^{C_n^1},g_{f_n}^{C_n^2}\}| n=1, \cdots, N\}$.
That is, for all $n$ in $1 \leq n \leq N$, there are $N$ positive glyph pairs in the minibatch.
Therefore for each glyph, remaining $2(N-1)$ glyphs are negative samples.
Our model $F$ maps every glyph images in the minibatch into font representation vectors in the latent space. 
The similarity of the embedding fonts for positive pairs and for negative pairs are defined by 
\begin{align}
\label{eq:sim-tau}
\texttt{pos-sim}({f}_n)&=\exp \left(\mathtt{sim}\left(   F(g_{f_n}^{C_n^1}), F(g_{f_n}^{C_n^2})\right) / \tau \right),\\
\texttt{neg-sim}_{k}({f}_n, {f}_{m \ne n}) &= \sum_{l=1}^{2}
\exp \left(\mathtt{sim}\left(  F(g_{f_n}^{C_n^k}), F(g_{f_m}^{C_m^l}) \right) / \tau \right), 
\end{align}
where $\tau$ is temperature scaling parameter.
Then  final loss $\mathcal{L}$ is sum of losses for each learning font $f_n$ is given by
\begin{equation}
\label{eq:pair-loss}
  \mathcal{L}=
  \frac{1}{N} \sum_{n=1}^{N} \left( -\sum_{k=1}^2 \log \frac{\texttt{pos-sim}({f}_n)}{\texttt{pos-sim}({f}_n)+\sum_{m=1, m \ne n}^{N} \texttt{neg-sim}_{k}({f}_n, {f}_{m \ne n})} \right).
\end{equation}
The loss~(\ref{eq:pair-loss}) is derived from  ``the normalized temperature-scaled cross entropy loss''~\cite{simclr}.

\section{Experiments}


\subsection{Baselines}

\begin{figure*}[]\centering
\includegraphics[width=1\linewidth]{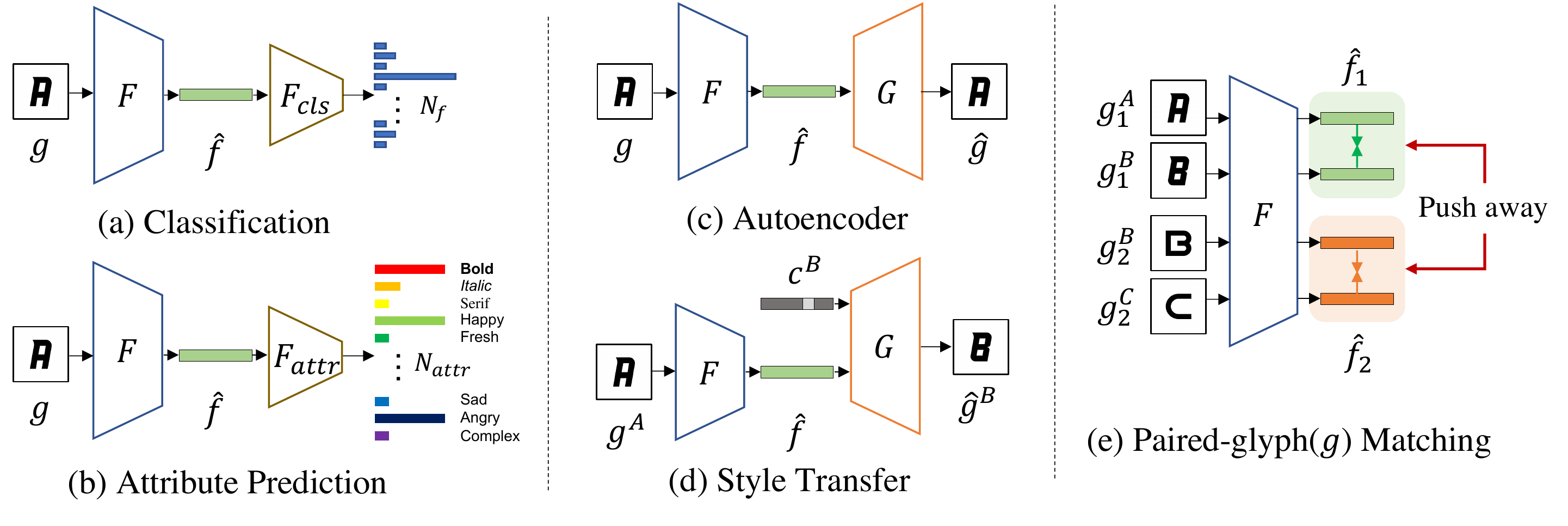}
\vspace{-0.5cm}
\caption{
Font representation learning techniques. $\hat{f}$ is a font embedding in the latent space for each method.
} 
\label{fig:font-embedding-baselines}
\end{figure*}

Figure~\ref{fig:font-embedding-baselines} shows baselines of font representation learning technique and our method {\it Paired-glyph Matching}.
These methods all share font embedding network $F$ as their backbone network.
We consider the output of $F$ from glyph $g$, $\hat{f}=F(g)$ as font embedding.
Comparing font representation learning baselines~(i.e., Classification~\cite{font-cls},  Style Transfer~\cite{DG-Font, zhang2018separating},  Autoencoder~\cite{autoenc1, autoenc2}, Attribute Prediction~\cite{odonovan:2014:font, Chen2019LargeScaleTF} and Srivatsan~\etal~\cite{srivatsan-etal-2019-deep}) are more described in Section~\ref{subsec:Baseline models} of the supplementary material.

\subsection{Datasets}
\label{sec:datasets}

O'Donovan~\etal~\cite{odonovan:2014:font} dataset contains $1,088$ fonts for the training set ($\mathbb{F}_\text{train}=\{f_i| 1 \leq i \leq 1,088\}$)  and $28$ fonts for the validation set.
Each font contains $62$ alphanumeric characters ($\mathbb{C}_\text{0-Z}$).
Thus, there are total $1,088 \times 62$ glyph images in training set.
Among the fonts in the training set, each font in $\{f_i| 1 \leq i \leq 120\}$ is annotated by $37$ attributes.
Each attribute is described by a  high-level expression, such as ``dramatic" or ``legible".
Each attribute value ranges from 0 to 1. The attribute value vector of each font in $\{f_i| 1 \leq i \leq 120\}$ is denoted by $a_i \in \mathbb{A}$, where $\mathbb{A}$ is the attribute set, i.e., $\mathbb{A} = \{a_1, a_2, ..., a_{120}\}, \forall a_i \in [0,1]^{37}$.
The remaining fonts of the training set (i.e., $\{f_i| 121 \leq i \leq 1,088\}$) are not annotated by any attributes. 


Open Font Library~(OFL), which is provided by Google Fonts\footnote{https://github.com/google/fonts}, provides $1,076$ typefaces (font families).
A typeface consists of several fonts that share a specific design.
In this paper, we do not consider typeface, thus, fonts in a  typeface are regarded as different fonts.
For instance, the typeface ``Bauer Bodoni'' includes  ``regular'', ``bold'', and ``italic'' fonts, which are considered  different fonts in our work.
Finally, we collected $3,802$ fonts for the alphanumeric character set ($\mathbb{C}_\text{0-Z}$).
We randomly partitioned $3,702$ fonts for the training set and the remaining $100$ fonts for the validation set.
Since these fonts are provided in  ``ttf'' and ``otf'' file formats, we  converted each font file into $62$ glyph images.

Capitals64~\cite{mcgan}, which was  used by Srivatsan~\etal~\cite{srivatsan-etal-2019-deep}, contains capital letters~($\mathbb{C}_\text{A-Z}$). 
The dataset is split into train, validation, and test sets of $7,649$, $1,473$, and $1560$ fonts, respectively.
We used this dataset to compare our method with Srivatsan~\etal method.

\subsection{Implementation Details}

Throughout all experiments, we used a single NVIDIA 2080ti or 1080ti gpu.
We did not observe a performance boost by tuning the last dimension of the projection head, as the previous research~\cite{simclr}.
Random sized crop augmentation was only used in our {\it Paired-glyph Matching} and {\it  Attribute Prediction}  as it degrades the  retrieval mean accuracy of other baselines.
The batch size was $64$ samples for each font, and the image input size was $64 \times 64$.
Bigger image size did not gain benefit on the retrieval mean accuracy score.
We used glyphs representing $\mathbb{C}_\text{a-Z}$ for the O'Donovan and OFL datasets and $\mathbb{C}_\text{A-Z}$ for the Capitals64 dataset.
We used the Adam~\cite{adam} optimizer with a learning rate of $2\mathrm{e}{-4}$ for all models and datasets.
We used ResNet18~\cite{resnet} as the backbone network of font embedding network $F$ for all models because other deeper neural network architectures were not effective.
Font embedding was average pooled vector from output of the backbone network.
The temperature scaling parameter $\tau$ of Equation~\ref{eq:sim-tau} has been used as $0.1$ for the OFL and Capitals64 dataset and $0.2$ for the O'Donovan dataset.

Denoting feature dimension by $\texttt{feat\_dim}$, we used $\texttt{feat\_dim}=512$ for all models for training the O'Donovan dataset and $\texttt{feat\_dim}=1,024$ for all models for the bigger OFL dataset.
We used $5$ transposed convolution layers and a last up-sample layer for the generator network $G$ of {\it  Autoencoder} and {\it  Style Transfer} models to generate $64 \times 64$ dimensional images from font embedding vectors.
The last 4 transposed convolutions were followed by self-attention modules~\cite{zhang2018rcan, Woo_2018_ECCV} and instance  normalization~\cite{ulyanov_instance_2016}.
The generator $G$ of {\it  Style Transfer} accepts ($\texttt{feat\_dim} + |\mathbb{C}|$)-dimensional vector, which is concatenation of font and one-hot character embedding.
Denoting a fully connected layer of the weight matrix $d_1 \times d_2$ as $\text{FC}^{(d_1 \times d_2)}$, the {\it  Classification} head ($F_{cls}$)is 
$\text{FC}^{(\texttt{font\_dim} \times |\mathbb{F}_\text{train}|)}$, 
the {\it  Attribute Prediction} head ($F_{attr}$) is 
$\text{FC}^{(\texttt{font\_dim} \times 37)}$.
Following previous research~\cite{simclr}, we also used a projection head and L2 feature normalization on {\it Paired-glyph Matching}.
The projection head is $\text{FC}^{(\texttt{font\_dim}  \times \texttt{font\_dim} )}-\text{ReLU}-\text{FC}^{(\texttt{font\_dim}  \times 70)}$.
More details (e.g., the generator $G$ architecture of {\it Style Transfer} and {\it Autoencoder} ) are presented in Section~\ref{sec:supp_implementation_details} of the supplementary material.
Codes are available at \url{https://github.com/junhocho/paired-glyph-matching}.

\subsection{Experimental Results}

To evaluate how well glyphs in a font are embedded in the latent space, we use the retrieval mean accuracy ($\underset{Ret}{\texttt{\textbf{MACC}}}(\mathbb{C}_\text{a-Z})$) as described in Section~\ref{subsec:Evaluation Metric} of the supplementary material.

\subsubsection{Evaluation on Unseen Fonts (O'Donovan and OFL datasets)}
\label{sec:Un-seen Fonts}

\begin{table}[t!]
\centering
\caption{
Performance evaluation ($\underset{Ret}{\texttt{\textbf{MACC}}}(\mathbb{C}_\text{a-Z})$ ) when trained on the O'Donovan dataset. 
}
\vspace{-0.15cm}
\label{tab:unseen-fonts-donovan}

\begin{tabular}{l|c|c|c}
\noalign{\smallskip}\noalign{\smallskip}\hline\hline
\multirow{2}{*}{Methods}                      
& \multirow{2}{*}{Data portion} 
& O'Donovan                                                 & OFL                                                        \\
                                       &                                                                                                          & $\underset{Ret}{\texttt{\textbf{MACC}}}(\mathbb{C}_\text{a-Z})$ & $\underset{Ret}{\texttt{\textbf{MACC}}}(\mathbb{C}_\text{a-Z})$  \\ 
\hline
\multirow{3}{*}{$\hat{f}$ of Paired-$g$ Matching} & $\{f_i| 1 \leq i \leq 1,088\}$ + $\mathbb{A}$                                                            & \textbf{89.91}                                                           & \textbf{66.46  }                                                          \\
                                       & $\{f_i| 1 \leq i \leq 1,088\}$                                                                           & \textbf{89.60 }                                                          & \textbf{64.53}                                                            \\
                                       & $\{f_i| 1 \leq i \leq 120\}$                                                                             & \textbf{72.03  }                                                         & \textbf{45.06}                                                            \\ 
\hline
\multirow{3}{*}{$\hat{f}$ of Classification~\cite{font-cls}}   & $\{f_i| 1 \leq i \leq 1,088\}$ + $\mathbb{A}$                                                            & 83.11                                                           & 58.56                                                            \\
                                       & $\{f_i| 1 \leq i \leq 1,088\}$                                                                           & 83.90                                                           & 57.36                                                            \\
                                       & $\{f_i| 1 \leq i \leq 120\}$                                                                             & 63.08                                                           & 35.33                                                            \\ 
\hline
\multirow{3}{*}{$\hat{f}$ of Style Transfer~\cite{DG-Font, zhang2018separating}}   & $\{f_i| 1 \leq i \leq 1,088\}$ + $\mathbb{A}$                                                            & 76.71                                                           & 36.71                                                            \\
                                       & $\{f_i| 1 \leq i \leq 1,088\}$                                                                           & 71.84                                                           & 36.88                                                            \\
                                       & $\{f_i| 1 \leq i \leq 120\}$                                                                             & 65.07                                                           & 30.00                                                            \\ 
\hline
\multirow{3}{*}{$\hat{f}$ of Autoencoder~\cite{autoenc1, autoenc2}}      & $\{f_i| 1 \leq i \leq 1,088\}$ + $\mathbb{A}$                                                            & 57.87                                                           & 31.97                                                            \\
                                       & $\{f_i| 1 \leq i \leq 1,088\}$                                                                           & 27.13                                                           & 13.96                                                            \\
                                       & $\{f_i| 1 \leq i \leq 120\}$                                                                             & 29.43                                                           & 12.31                                                            \\ 
\hline
$\hat{f}$ of Attribute Pred.~\cite{odonovan:2014:font, Chen2019LargeScaleTF}              & $\{f_i| 1 \leq i \leq 120\}$ + $\mathbb{A}$                                                              & 64.08                                                           & 38.02                                                            \\
\hline
\hline
\end{tabular}

\end{table}

\begin{table}[t!]
\centering
\caption{Performance evaluation ($\underset{Ret}{\texttt{\textbf{MACC}}}(\mathbb{C}_\text{a-Z})$ ) when trained on the OFL dataset. Paired-glyph Matching $\dagger$ and $\ddagger$ are each  trained with different similarity-based losses~\cite{schroff2015facenet, huang2020pica}.}
\vspace{-0.15cm}
\label{tab:unseen-fonts-ofl}
\begin{tabular}{c|c|c}
\noalign{\smallskip}\noalign{\smallskip}\hline\hline
Methods &
OFL valset  &
O'Donovan valset  \\


\hline

 
\rule{0pt}{11pt} 
$\hat{f}$ of Paired-glyph Matching & {\bf 91.82} & {\bf 75.44} \\
$\hat{f}$ of Classification~\cite{font-cls}  & 83.67 & 68.48 \\
$\hat{f}$ of Style Transfer~\cite{DG-Font, zhang2018separating} & 82.24 & 46.23 \\
$\hat{f}$ of Autoencoder~\cite{autoenc1, autoenc2} & 15.55 & 26.66 \\
\hline
$\hat{f}$ of Paired-glyph Matching $\dagger$& 88.93 & 	72.98 \\
$\hat{f}$ of Paired-glyph Matching $\ddagger$&	82.70 &	54.28 \\

\hline
\hline

 
 

\end{tabular}
\vspace{-0.5cm}
\end{table}

Table~\ref{tab:unseen-fonts-donovan} presents the performances on font embeddings $\hat{f}$ of all  methods~(i.e., {\it Paired-glyph Matching,  Classification,  Style Transfer,  Autoencoder} and {\it  Attribute Prediction}) depending on training data portion in the O'Donovan dataset.
For every $100$ epochs until $15,000$ epochs, we evaluated the models on the O'Donovan validation set with the retrieval mean accuracy.
Note that models had not seen fonts in the validation set.
We found and reported the best score on the O'Donovan validation set and then evaluated the model with same weights on the OFL validation set.
First of all, we compare when training only small portion ($\{f_i | 1 \leq i \leq 120\}$) of fonts in Table~\ref{tab:unseen-fonts-donovan} and the performance was excellent in the order of {\it Paired-glyph Matching} (72.03), {\it  Style Transfer} (65.07), {\it  Attribute Prediction}  (64.08), {\it  Classification} (63.08), {\it  Autoencoder} (29.43).
It is notable that {\it Paired-glyph Matching} outperformed {\it  Attribute Prediction} even without richer font annotations $\mathbb{A}$.
To see the effectiveness of font attribute data $\mathbb{A}$, we jointly trained {\it Paired-glyph Matching,  Classification,  Style Transfer,  Autoencoder} with {\it  Attribute Prediction} ($F_{attr}$ in Figure~\ref{fig:font-embedding-baselines} (b)) and reported as $\{f_i | 1 \leq i \leq 120\} +\mathbb{A}$ in the data portion column of Table~\ref{tab:unseen-fonts-donovan}.
We found training font attributes ($+\mathbb{A}$) to have no significant difference in {\it Paired-glyph Matching} and {\it Classification}.
This indicates that font attribute data may not be worth the high annotation cost to train font representations.

Table~\ref{tab:unseen-fonts-ofl} presents the performances of all font embedding methods trained on the OFL dataset. We found and reported the best score on the OFL validation set until $25,000$ epochs and then evaluated the model with same weights on the O'Donovan validation set. 
Since, there are more possible solutions (e.g, triplet loss~\cite{schroff2015facenet} or other self-supervised methods~\cite{ huang2020pica, contrastive-learning2, contrastive-learning3}) to learn similarities in paired-glyph matching learning, we include Paired Glyph Matching $\dagger$,  $\ddagger$ which are respectively trained with losses based on  deep clustering algorithm (PICA)~\cite{huang2020pica} and triplet loss~\cite{schroff2015facenet}.
We observed that paired-glyph matching learning with the loss~(\ref{eq:pair-loss}) performed the best compare to other similarity learning approaches $\dagger$,  $\ddagger$~\cite{huang2020pica, schroff2015facenet}.

\begin{figure}[]
  \centering
  \includegraphics[width=0.855\linewidth]{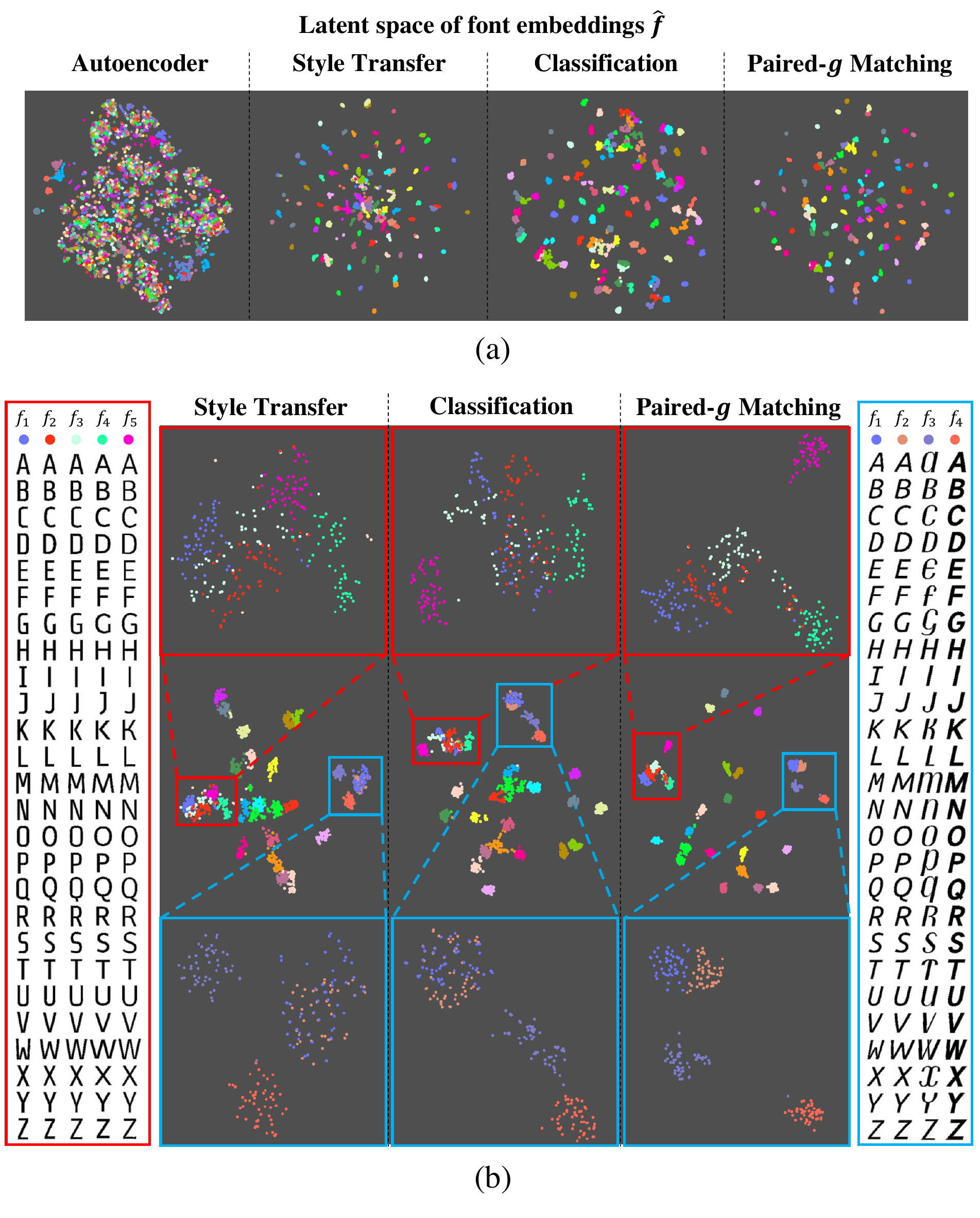}
  \vspace{-0.5cm}
  \caption{
  (a) Font latent space of (a) the OFL dataset and (b) the O'Donovan dataset annotated by font classes for each embedding methods. The {\color{red}Red} and the {\color{cyan}cyan} boxes respectively include font classes $\{f_1, f_2, f_3, f_4, f_5\}$ and $\{f_6, f_7, f_8, f_9\}$.
  }
  \label{fig:donovan-font-embeddings}
  \vspace{-0.7cm}
  
\end{figure}

To visually understand how comparing methods perform, we used T-SNE~\cite{tsne} projection on the font latent space as in Figure~\ref{fig:donovan-font-embeddings}.
From observations in the font latent space of the OFL dataset (Figure~\ref{fig:donovan-font-embeddings} (a)) and the O'Donovan dataset (Figure~\ref{fig:donovan-font-embeddings} (b)), glyphs in a font were better clustered in the order of {\it Paired-glyph Matching}, {\it  Classification}, and {\it  Style Transfer}.
In particular, note the \textcolor{red}{red} and \textcolor{cyan}{cyan} boxes in Figure~\ref{fig:donovan-font-embeddings} (b).
{\it Style Transfer} and {\it Classification} methods do not distinguish the glyphs of the fonts $f_1, f_2, f_3$ in the \textcolor{red}{red box} and $f_6$, $f_7$ in the \textcolor{cyan}{cyan box}, but our method distinguished them relatively well.

\subsubsection{Evaluation on Unseen Fonts (Capitals64 dataset)}

Table~\ref{tab:unseen-fonts-capitals64} presents the performances of font representation learning  methods~(i.e., {\it Paired-glyph Matching,  Classification,  Style Transfer,  Autoencoder} and Srivatsan~\etal~\cite{srivatsan-etal-2019-deep}) on the Capitals64 dataset and the O'Donovan dataset.
Similar to Table~\ref{tab:unseen-fonts-donovan} and \ref{tab:unseen-fonts-ofl}, our method performs the best in the retrieval mean accuracy $\underset{Ret}{\texttt{\textbf{MACC}}}(\mathbb{C}_\text{A-Z})$ measure.
To more quantitatively evaluate representation power of $\hat{f}$, we trained font attribute ($\mathbb{A}$) prediction task, which is similar to  linear evaluation protocol~\cite{simclr}.
That is, we train a linear mapping from font embedding $\hat{f}$ of each method to  $37$ font attributes and validate with $L_1$-prediction error.
We trained 120 fonts and validated 28 fonts in O'Donovan dataset, varying learning rate in range of $[  1\mathrm{e}{-6}, 1\mathrm{e}{-5}, 1\mathrm{e}{-4}, 1\mathrm{e}{-3}, 1\mathrm{e}{-2}]$ and reported the lowest $L_1$-error in Table~\ref{tab:unseen-fonts-capitals64} last column.
Our method outperformed Srivatsan~\etal by predicting font attributes with lower error.

\begin{figure*}[]\centering
\includegraphics[width=1.\linewidth]{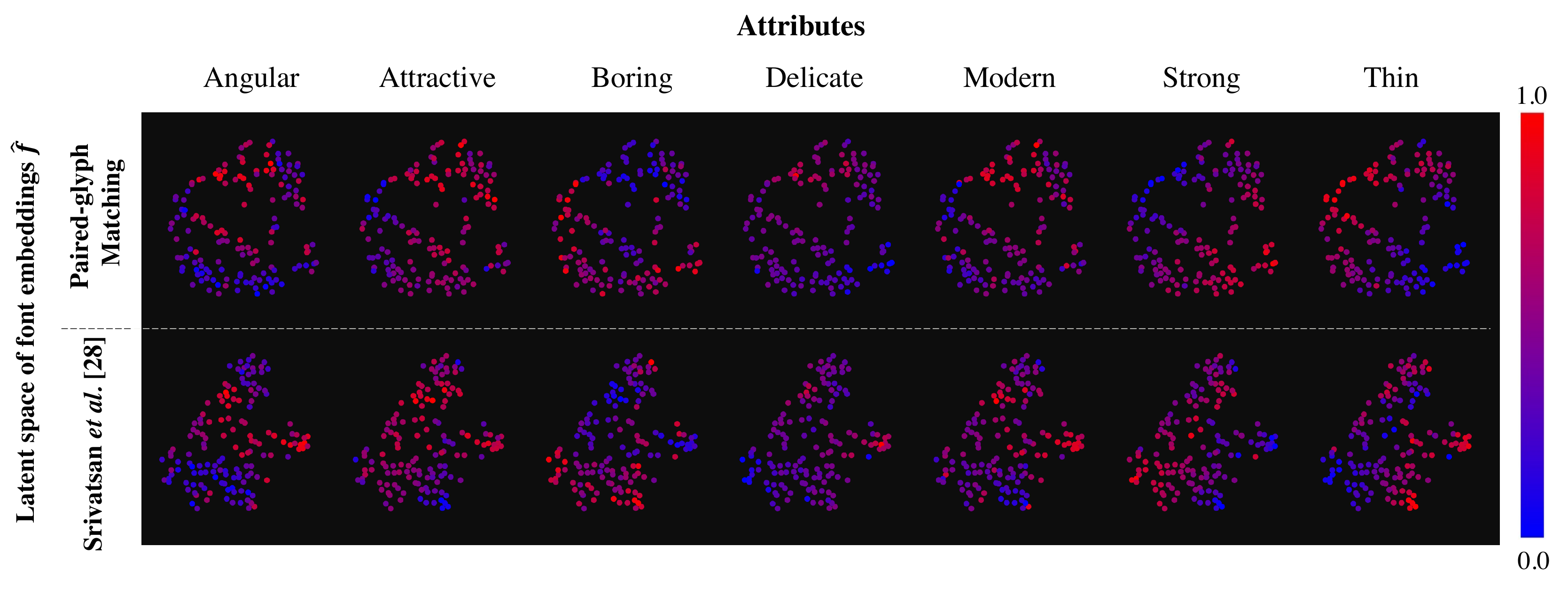}
\vspace*{-0.5cm}
\caption{Font latent space of {\it Paired-glyph Matching} and Srivatsan~\etal~\cite{srivatsan-etal-2019-deep} annotated by font attributes~(i.e., Angular, Attractive, Boring, Delicate, Modern, Strong, Thin). Both methods are trained on Capitals64 and tested on the O'Donovan dataset.
} 
\label{fig:donovan-attribute-embeddings}
\vspace{-0.5cm}
\end{figure*}

In Figure~\ref{fig:donovan-attribute-embeddings}, we observed the latent space of the O'Donovan fonts with attribute annotations  $\mathbb{A}$.
Refer to Srivatsan~\etal, we took max-pooling operation on embeddings of glyphs in a font and regarded it as the font embedding.
Each font in the O'Donovan dataset is colored with respective attribute value in Figure~\ref{fig:donovan-attribute-embeddings}.
Despite not training on font attribute data ($\mathbb{A}$), both methods gathered fonts according to values of the font attributes.

\begin{table*}[t!]
\centering
\caption{Performance evaluation ($\underset{Ret}{\texttt{\textbf{MACC}}}(\mathbb{C}_\text{A-Z})$ and Font attribute prediction) on the Capitals64 dataset. All methods are trained on Capitals64, validated and tested on Capitals64. Font attribute prediction is evaluated on O'Donovan dataset  with $L_1$-error.}
\label{tab:unseen-fonts-capitals64}
\begin{tabular}{c|c|c|c}
\noalign{\smallskip}\noalign{\smallskip}\hline\hline
 

 \multirow{2}{*}{
        Methods
        }
        
      & Captials64 valset
      & Captials64 testset 
      & O'Donovan 
      \\
      & $\underset{Ret}{\texttt{\textbf{MACC}}}(\mathbb{C}_\text{A-Z})$
      & $\underset{Ret}{\texttt{\textbf{MACC}}}(\mathbb{C}_\text{A-Z})$
      & $L_1$-error
      \\                  
\hline
 
\rule{0pt}{11pt} 
$\hat{f}$ of Paired-$g$ Matching & {\bf 61.38} & {\bf 62.66} & {\bf 0.09589
}\\
$\hat{f}$ of Classification~\cite{font-cls}  & 55.27 & 56.31 & 0.1275 \\
$\hat{f}$ of Style Transfer~\cite{DG-Font, zhang2018separating} & 32.22 & 32.53 & 0.1217 \\
$\hat{f}$ of Autoencoder~\cite{autoenc1, autoenc2} & 13.60 & 14.16 & 0.1312 \\
$\hat{f}$ of Srivatsan~\etal ~\cite{srivatsan-etal-2019-deep} & 11.72 & 11.56 & 0.1097
 \\ 
\hline
\hline

\end{tabular}
\end{table*}

\subsubsection{Transfer Learning to Font Style Transfer \& Generation.}
\label{sec:transfer-learning-font-style-transfer}

\begin{figure}[]
  \centering
\includegraphics[width=1\linewidth]{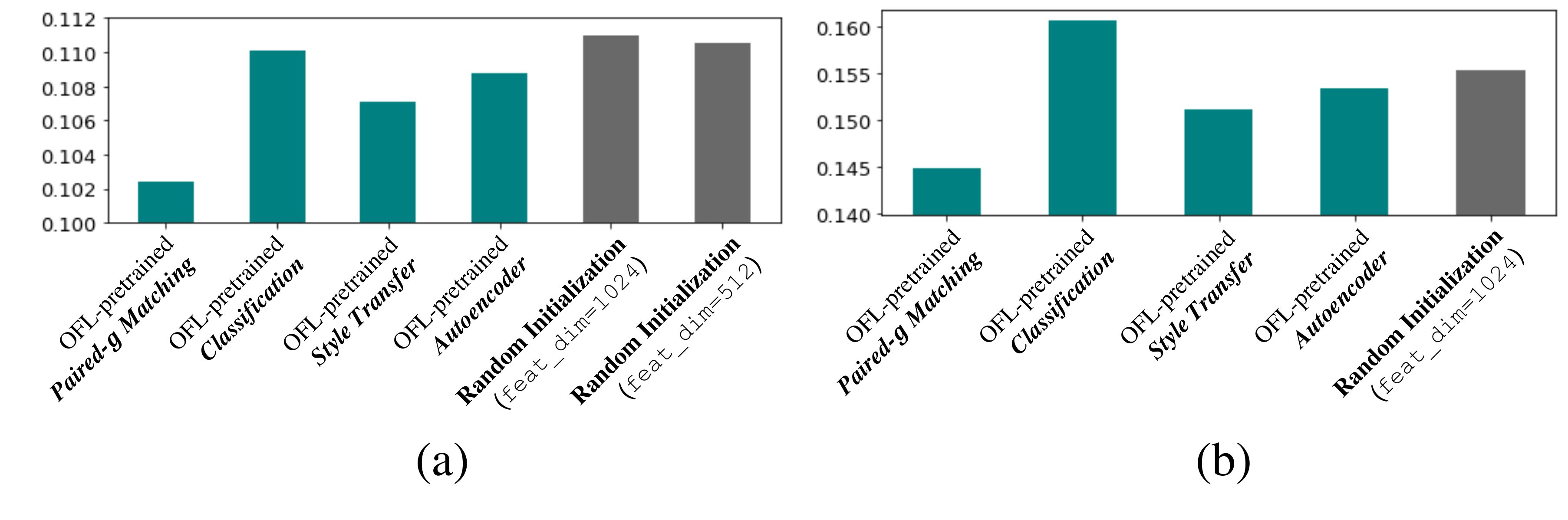}
  
  \caption{
  Image $\texttt{L1-error}$ measured in (a) font style transfer and (b) font generation (Attr2Font~\cite{WangSIGGRAPH2020})  with the \textcolor{teal}{OFL dataset pretrained} and \textcolor{gray}{random} initialization.
  }
  \label{fig:downstream-font-style-transfer-bar-graph}

\vspace{-0.5cm}
\end{figure}

In this experiment, we checked the transfer learning performance in font style transfer (See Section~\ref{sec:style-transfer}) and font generation~(Attr2Font~\cite{WangSIGGRAPH2020}) as downstream tasks.
We used pretrained weights from the best-performing models $F$ (i.e., {\it Paired-glyph Matching,  Classification,  Style Transfer} and {\it  Autoencoder} from Table~\ref{tab:unseen-fonts-ofl}) on the OFL dataset and applied transfer learning to O'Donovan dataset, which is smaller the OFL dataset.
To evaluate the generation quality of font style transfer model, we calculated average $L_1$ errors for all images generated from a input glyph and an one-hot character embedding as follows:
\begin{equation}
    \texttt{L1-error} = \dfrac{1}{|\mathbb{F}_\text{val}| \times I_\text{dim} }\sum_{f \in \mathbb{F}_\text{val} } \sum_
    {\text{C}_i , \text{C}_j \in \mathbb{C}  }  \|G\left(F(g^{\text{C}_i}_f), c^{\text{C}_j}\right) - g^{\text{C}_j}_f)\|_1, \nonumber
\end{equation}
where $I_\text{dim}=H \times W \times C$ is number of pixels in an image.
For the Attr2Font model~\cite{WangSIGGRAPH2020}, which performs  attribute-based font generation as a downstream task, we initialized the ``style encoder''  with the aforementioned pretrained weights, and \texttt{L1-error} is similarly defined.
Note that we scratch-train the generator $G$ weights of {\it  Autoencoder} and {\it  Style Transfer}.
In Figures~\ref{fig:downstream-font-style-transfer-bar-graph},
we measured performance gains of \textcolor{teal}{pretrained models} over \textcolor{gray}{random initialized baseline}.
Interestingly, the models trained in the generative way (i.e., {\it  Autoencoder,  Style Transfer}) on the OFL dataset seemed to be better in the downstream generative tasks than the model trained through {\it Classification}.
As a result, we determined that {\it Paired-glyph Matching} performed the best, showing that our method can be useful as transfer learning to the generative tasks.

\section{Conclusion}

In this paper, we proposed a new discriminative font embedding method that attracts the representations of glyphs in the same font to one another but pushes away glyphs in other fonts.
Our method needed neither a generator network nor font attribute tags because we actively take advantage of {\it Glyph-font-consistency}.
Through extensive evaluation, we show our model outperformed the conventional representation learning techniques for generalization to unseen fonts.
Finally, we confirmed the benefits of our method for transfer learning in the font style transfer and generation tasks.

\section*{Acknowledgement}

This work was partly supported by Institute of Information \& communications Technology Planning \& Evaluation (IITP) grant funded by the Korea government (MSIT) [NO.2021-0-01343, Artificial Intelligence Graduate School Program (Seoul National University)] and IITP grant funded by Korea government(MSIT) [No.B0101-15-0266, Development of High Performance Visual BigData Discovery Platform for Large-Scale Realtime Data Analysis].

\bibliography{egbib}

\begin{thebibliography}{45}
\providecommand{\natexlab}[1]{#1}
\providecommand{\url}[1]{\texttt{#1}}
\expandafter\ifx\csname urlstyle\endcsname\relax
  \providecommand{\doi}[1]{doi: #1}\else
  \providecommand{\doi}{doi: \begingroup \urlstyle{rm}\Url}\fi

\bibitem[Azadi et~al.(2018)Azadi, Fisher, Kim, Wang, Shechtman, and
  Darrell]{mcgan}
Samaneh Azadi, Matthew Fisher, Vladimir Kim, Zhaowen Wang, Eli Shechtman, and
  Trevor Darrell.
\newblock Multi-content gan for few-shot font style transfer.
\newblock In \emph{Proceedings of the IEEE Conference on Computer Vision and
  Pattern Recognition}, volume~11, page~13, 2018.

\bibitem[Baird and Nagy(1994)]{boost-text-recog2}
Henry~S. Baird and George Nagy.
\newblock {Self-correcting 100-font classifier}.
\newblock In Luc~M. Vincent and Theo Pavlidis, editors, \emph{Document
  Recognition}, volume 2181, pages 106 -- 115. International Society for Optics
  and Photonics, SPIE, 1994.
\newblock \doi{10.1117/12.171098}.
\newblock URL \url{https://doi.org/10.1117/12.171098}.

\bibitem[Ben~Moussa et~al.(2010)Ben~Moussa, Zahour, Benabdelhafid, and
  Alimi]{fontcls3}
Sami Ben~Moussa, Abderrazak Zahour, Abdellatif Benabdelhafid, and Adel Alimi.
\newblock New features using fractal multi-dimensions for generalized arabic
  font recognition.
\newblock \emph{Pattern Recognition Letters}, 31:\penalty0 361--371, 04 2010.
\newblock \doi{10.1016/j.patrec.2009.10.015}.

\bibitem[Chen et~al.(2019)Chen, Wang, Xu, Jin, and Luo]{Chen2019LargeScaleTF}
Tianlang Chen, Zhaowen Wang, N.~Xu, Hailin Jin, and Jiebo Luo.
\newblock Large-scale tag-based font retrieval with generative feature
  learning.
\newblock \emph{2019 IEEE/CVF International Conference on Computer Vision
  (ICCV)}, pages 9115--9124, 2019.

\bibitem[Chen et~al.(2020)Chen, Kornblith, Norouzi, and Hinton]{simclr}
Ting Chen, Simon Kornblith, Mohammad Norouzi, and Geoffrey Hinton.
\newblock A simple framework for contrastive learning of visual
  representations.
\newblock \emph{arXiv preprint arXiv:2002.05709}, 2020.

\bibitem[Dosovitskiy et~al.(2014)Dosovitskiy, Springenberg, Riedmiller, and
  Brox]{contrastive-learning2}
Alexey Dosovitskiy, Jost~Tobias Springenberg, Martin Riedmiller, and Thomas
  Brox.
\newblock Discriminative unsupervised feature learning with convolutional
  neural networks.
\newblock \emph{Advances in neural information processing systems}, 27, 2014.

\bibitem[Friedman(2001)]{10.1214/aos/1013203451}
Jerome~H. Friedman.
\newblock {Greedy function approximation: A gradient boosting machine.}
\newblock \emph{The Annals of Statistics}, 29\penalty0 (5):\penalty0 1189 --
  1232, 2001.
\newblock \doi{10.1214/aos/1013203451}.
\newblock URL \url{https://doi.org/10.1214/aos/1013203451}.

\bibitem[Geirhos et~al.(2019)Geirhos, Rubisch, Michaelis, Bethge, Wichmann, and
  Brendel]{geirhos2018imagenettrained}
Robert Geirhos, Patricia Rubisch, Claudio Michaelis, Matthias Bethge, Felix~A.
  Wichmann, and Wieland Brendel.
\newblock Imagenet-trained {CNN}s are biased towards texture; increasing shape
  bias improves accuracy and robustness.
\newblock In \emph{International Conference on Learning Representations}, 2019.
\newblock URL \url{https://openreview.net/forum?id=Bygh9j09KX}.

\bibitem[Goodfellow et~al.(2014)Goodfellow, Pouget-Abadie, Mirza, Xu,
  Warde-Farley, Ozair, Courville, and Bengio]{gan_ian}
Ian Goodfellow, Jean Pouget-Abadie, Mehdi Mirza, Bing Xu, David Warde-Farley,
  Sherjil Ozair, Aaron Courville, and Yoshua Bengio.
\newblock Generative adversarial nets.
\newblock In Z.~Ghahramani, M.~Welling, C.~Cortes, N.~Lawrence, and K.~Q.
  Weinberger, editors, \emph{Advances in Neural Information Processing
  Systems}, volume~27. Curran Associates, Inc., 2014.
\newblock URL
  \url{https://proceedings.neurips.cc/paper/2014/file/5ca3e9b122f61f8f06494c97b1afccf3-Paper.pdf}.

\bibitem[Hassan et~al.(2021)Hassan, Ahmed, and Choi]{Hassan2021UnpairedFF}
Ammar~Ul Hassan, Hammad Ahmed, and Jaeyoung Choi.
\newblock Unpaired font family synthesis using conditional generative
  adversarial networks.
\newblock \emph{Knowl. Based Syst.}, 229:\penalty0 107304, 2021.

\bibitem[Hayashi et~al.(2019)Hayashi, Abe, and Uchida]{Hayashi2019GlyphGANSF}
Hideaki Hayashi, Kohtaro Abe, and Seiichi Uchida.
\newblock Glyphgan: Style-consistent font generation based on generative
  adversarial networks.
\newblock \emph{ArXiv}, abs/1905.12502, 2019.

\bibitem[He et~al.(2015)He, Zhang, Ren, and Sun]{resnet}
Kaiming He, Xiangyu Zhang, Shaoqing Ren, and Jian Sun.
\newblock Deep residual learning for image recognition.
\newblock \emph{arXiv preprint arXiv:1512.03385}, 2015.

\bibitem[Isola et~al.(2017)Isola, Zhu, Zhou, and Efros]{pix2pix2017}
Phillip Isola, Jun-Yan Zhu, Tinghui Zhou, and Alexei~A Efros.
\newblock Image-to-image translation with conditional adversarial networks.
\newblock \emph{CVPR}, 2017.

\bibitem[Jiabo~Huang and Zhu(2020)]{huang2020pica}
Shaogang~Gong Jiabo~Huang and Xiatian Zhu.
\newblock Deep semantic clustering by partition confidence maximisation.
\newblock In \emph{Proceedings of IEEE Conference on Computer Vision and
  Pattern Recognition (CVPR)}, 2020.

\bibitem[Kang et~al.(2021)Kang, Haraguchi, Kimura, and Uchida]{kang2021shared}
Jihun Kang, Daichi Haraguchi, Akisato Kimura, and Seiichi Uchida.
\newblock Shared latent space of font shapes and impressions.
\newblock \emph{arXiv preprint arXiv:2103.12347}, 2021.

\bibitem[Kataria et~al.(2010)Kataria, Marchesotti, and
  Perronnin]{font-retrieval-icip2010}
Saurabh Kataria, Luca Marchesotti, and Florent Perronnin.
\newblock Font retrieval on a large scale: An experimental study.
\newblock In \emph{2010 IEEE International Conference on Image Processing},
  pages 2177--2180, 2010.
\newblock \doi{10.1109/ICIP.2010.5650155}.

\bibitem[Kingma and Ba(2014)]{adam}
Diederik Kingma and Jimmy Ba.
\newblock Adam: A method for stochastic optimization.
\newblock \emph{International Conference on Learning Representations}, 12 2014.

\bibitem[Kingma and Welling(2013)]{vae}
Diederik~P Kingma and Max Welling.
\newblock Auto-encoding variational bayes.
\newblock \emph{arXiv preprint arXiv:1312.6114}, 2013.

\bibitem[Kulahcioglu and de~Melo(2020)]{fonts-like-this-happier-acmmm2020}
Tugba Kulahcioglu and Gerard de~Melo.
\newblock \emph{Fonts Like This but Happier: A New Way to Discover Fonts}, page
  2973–2981.
\newblock Association for Computing Machinery, New York, NY, USA, 2020.
\newblock ISBN 9781450379885.
\newblock URL \url{https://doi.org/10.1145/3394171.3413534}.

\bibitem[Lowe(1999)]{SIFT}
D.G. Lowe.
\newblock Object recognition from local scale-invariant features.
\newblock In \emph{Proceedings of the Seventh IEEE International Conference on
  Computer Vision}, volume~2, pages 1150--1157 vol.2, 1999.
\newblock \doi{10.1109/ICCV.1999.790410}.

\bibitem[Marcelino(2018)]{marcelino2018transfer}
Pedro Marcelino.
\newblock Transfer learning from pre-trained models.
\newblock \emph{Towards Data Science}, 10:\penalty0 23, 2018.

\bibitem[Mikolov et~al.(2013)Mikolov, Sutskever, Chen, Corrado, and
  Dean]{word2vec}
Tomas Mikolov, Ilya Sutskever, Kai Chen, Greg~S Corrado, and Jeff Dean.
\newblock Distributed representations of words and phrases and their
  compositionality.
\newblock In C.~J.~C. Burges, L.~Bottou, M.~Welling, Z.~Ghahramani, and K.~Q.
  Weinberger, editors, \emph{Advances in Neural Information Processing
  Systems}, volume~26. Curran Associates, Inc., 2013.
\newblock URL
  \url{https://proceedings.neurips.cc/paper/2013/file/9aa42b31882ec039965f3c4923ce901b-Paper.pdf}.

\bibitem[Odena et~al.(2017)Odena, Olah, and Shlens]{acgan}
Augustus Odena, Christopher Olah, and Jonathon Shlens.
\newblock Conditional image synthesis with auxiliary classifier gans.
\newblock In \emph{Proceedings of the 34th International Conference on Machine
  Learning - Volume 70}, ICML'17, page 2642–2651. JMLR.org, 2017.

\bibitem[O'Donovan et~al.(2014)O'Donovan, Libeks, Agarwala, and
  Hertzmann]{odonovan:2014:font}
Peter O'Donovan, Janis Libeks, Aseem Agarwala, and Aaron Hertzmann.
\newblock Exploratory {F}ont {S}election {U}sing {C}rowdsourced {A}ttributes.
\newblock \emph{ACM Transactions on Graphics (Proc. SIGGRAPH)}, 33\penalty0
  (4), 2014.

\bibitem[Pengcheng et~al.(2017)Pengcheng, Gang, Jiangqin, and
  Baogang]{fontcls5}
Gao Pengcheng, Gu~Gang, Wu~Jiangqin, and Wei Baogang.
\newblock Chinese calligraphic style representation for recognition.
\newblock \emph{Int. J. Doc. Anal. Recognit.}, 20\penalty0 (1):\penalty0
  59–68, mar 2017.
\newblock ISSN 1433-2833.
\newblock \doi{10.1007/s10032-016-0277-z}.
\newblock URL \url{https://doi.org/10.1007/s10032-016-0277-z}.

\bibitem[Schroff et~al.(2015)Schroff, Kalenichenko, and
  Philbin]{schroff2015facenet}
Florian Schroff, Dmitry Kalenichenko, and James Philbin.
\newblock Facenet: A unified embedding for face recognition and clustering.
\newblock In \emph{Proceedings of the IEEE conference on computer vision and
  pattern recognition}, pages 815--823, 2015.

\bibitem[Shi and Pavlidis(1997)]{boost-text-recog}
Hongwei Shi and T.~Pavlidis.
\newblock Font recognition and contextual processing for more accurate text
  recognition.
\newblock In \emph{Proceedings of the Fourth International Conference on
  Document Analysis and Recognition}, volume~1, pages 39--44 vol.1, 1997.
\newblock \doi{10.1109/ICDAR.1997.619810}.

\bibitem[Srivatsan et~al.(2019)Srivatsan, Barron, Klein, and
  Berg-Kirkpatrick]{srivatsan-etal-2019-deep}
Nikita Srivatsan, Jonathan Barron, Dan Klein, and Taylor Berg-Kirkpatrick.
\newblock A deep factorization of style and structure in fonts.
\newblock In \emph{Proceedings of the 2019 Conference on Empirical Methods in
  Natural Language Processing and the 9th International Joint Conference on
  Natural Language Processing (EMNLP-IJCNLP)}, pages 2195--2205, Hong Kong,
  China, November 2019. Association for Computational Linguistics.
\newblock \doi{10.18653/v1/D19-1225}.
\newblock URL \url{https://aclanthology.org/D19-1225}.

\bibitem[Taigman et~al.(2016)Taigman, Polyak, and Wolf]{dtn}
Yaniv Taigman, Adam Polyak, and Lior Wolf.
\newblock Unsupervised cross-domain image generation.
\newblock \emph{arXiv preprint arXiv:1611.02200}, 2016.

\bibitem[Tang et~al.(2020)Tang, Zhang, Chen, Wang, and Chen]{autoenc1}
Shancheng Tang, Puyue Zhang, Xiongxiong Chen, Hanbo Wang, and Ming Chen.
\newblock A word representation method based on glyph of chinese character.
\newblock In \emph{2020 International Conference on Intelligent Transportation,
  Big Data Smart City (ICITBS)}, pages 954--957, 2020.
\newblock \doi{10.1109/ICITBS49701.2020.00212}.

\bibitem[Tao et~al.(2016)Tao, Lin, Jin, and Li]{fontcls4}
Dapeng Tao, Xu~Lin, Lianwen Jin, and Xuelong Li.
\newblock Principal component 2-d long short-term memory for font recognition
  on single chinese characters.
\newblock \emph{IEEE Transactions on Cybernetics}, 46\penalty0 (3):\penalty0
  756--765, 2016.
\newblock \doi{10.1109/TCYB.2015.2414920}.

\bibitem[Tensmeyer et~al.(2017)Tensmeyer, Saunders, and Martinez]{font-cls}
Chris Tensmeyer, Daniel Saunders, and Tony Martinez.
\newblock Convolutional neural networks for font classification.
\newblock In \emph{2017 14th IAPR International Conference on Document Analysis
  and Recognition (ICDAR)}, volume~01, pages 985--990, 2017.
\newblock \doi{10.1109/ICDAR.2017.164}.

\bibitem[Tian(2017)]{zi2zi}
Yuchen Tian, Apr 2017.
\newblock URL \url{https://kaonashi-tyc.github.io/2017/04/06/zi2zi.html}.

\bibitem[Ulyanov et~al.(2016)Ulyanov, Vedaldi, and
  Lempitsky]{ulyanov_instance_2016}
Dmitry Ulyanov, Andrea Vedaldi, and Victor Lempitsky.
\newblock Instance {Normalization}: {The} {Missing} {Ingredient} for {Fast}
  {Stylization}.
\newblock \emph{arXiv:1607.08022 [cs]}, July 2016.
\newblock URL \url{http://arxiv.org/abs/1607.08022}.
\newblock arXiv: 1607.08022.

\bibitem[Van~den Oord et~al.(2018)Van~den Oord, Li, and
  Vinyals]{contrastive-learning3}
Aaron Van~den Oord, Yazhe Li, and Oriol Vinyals.
\newblock Representation learning with contrastive predictive coding.
\newblock \emph{arXiv e-prints}, pages arXiv--1807, 2018.

\bibitem[van~der Maaten and Hinton(2008)]{tsne}
Laurens van~der Maaten and Geoffrey Hinton.
\newblock Visualizing data using {t-SNE}.
\newblock \emph{Journal of Machine Learning Research}, 9:\penalty0 2579--2605,
  2008.
\newblock URL \url{http://www.jmlr.org/papers/v9/vandermaaten08a.html}.

\bibitem[Wang et~al.(2022)Wang, Zhu, Shen, Wang, Wu, and Yao]{autoenc2}
Chen Wang, Yani Zhu, Zhangyi Shen, Dong Wang, Guohua Wu, and Ye~Yao.
\newblock Font transfer based on parallel auto-encoder for glyph perturbation
  via strokes moving.
\newblock In Yongxuan Lai, Tian Wang, Min Jiang, Guangquan Xu, Wei Liang, and
  Aniello Castiglione, editors, \emph{Algorithms and Architectures for Parallel
  Processing}, pages 586--602, Cham, 2022. Springer International Publishing.
\newblock ISBN 978-3-030-95388-1.

\bibitem[Woo et~al.(2018)Woo, Park, Lee, and Kweon]{Woo_2018_ECCV}
Sanghyun Woo, Jongchan Park, Joon-Young Lee, and In~So Kweon.
\newblock Cbam: Convolutional block attention module.
\newblock In \emph{Proceedings of the European Conference on Computer Vision
  (ECCV)}, September 2018.

\bibitem[Xi et~al.(2020)Xi, Yan, Hua, and Zhong]{Xi2020JointFontGANJG}
Yankun Xi, Guoli Yan, Jing Hua, and Zichun Zhong.
\newblock Jointfontgan: Joint geometry-content gan for font generation via
  few-shot learning.
\newblock \emph{Proceedings of the 28th ACM International Conference on
  Multimedia}, 2020.

\bibitem[Xie et~al.(2021)Xie, Chen, Sun, and Lu]{DG-Font}
Yangchen Xie, Xinyuan Chen, Li~Sun, and Yue Lu.
\newblock Dg-font: Deformable generative networks for unsupervised font
  generation.
\newblock In \emph{Proceedings of the IEEE Conference on Computer Vision and
  Pattern Recognition}, 2021.

\bibitem[Yizhi~Wang*(2020)]{WangSIGGRAPH2020}
Zhouhui~Lian Yizhi~Wang*, Yue~Gao*.
\newblock Attribute2font: Creating fonts you want from attributes.
\newblock \emph{ACM Trans. Graph.}, 2020.

\bibitem[Zhang et~al.(2018{\natexlab{a}})Zhang, Zhang, and
  Cai]{zhang2018separating}
Yexun Zhang, Ya~Zhang, and Wenbin Cai.
\newblock Separating style and content for generalized style transfer.
\newblock In \emph{Proceedings of the IEEE Conference on Computer Vision and
  Pattern Recognition}, volume~1, 2018{\natexlab{a}}.

\bibitem[Zhang et~al.(2018{\natexlab{b}})Zhang, Li, Li, Wang, Zhong, and
  Fu]{zhang2018rcan}
Yulun Zhang, Kunpeng Li, Kai Li, Lichen Wang, Bineng Zhong, and Yun Fu.
\newblock Image super-resolution using very deep residual channel attention
  networks.
\newblock In \emph{ECCV}, 2018{\natexlab{b}}.

\bibitem[Zhu et~al.(2001)Zhu, Tan, and Wang]{fontcls2}
Yong Zhu, Tieniu Tan, and Yunhong Wang.
\newblock Font recognition based on global texture analysis.
\newblock \emph{IEEE Transactions on Pattern Analysis and Machine
  Intelligence}, 23\penalty0 (10):\penalty0 1192--1200, 2001.
\newblock \doi{10.1109/34.954608}.

\bibitem[Zramdini and Ingold(1998)]{fontcls1}
Abdelwahab Zramdini and Rolf Ingold.
\newblock Optical font recognition using typographical features.
\newblock \emph{IEEE Trans. Pattern Anal. Mach. Intell.}, 20\penalty0
  (8):\penalty0 877–882, aug 1998.
\newblock ISSN 0162-8828.
\newblock \doi{10.1109/34.709616}.
\newblock URL \url{https://doi.org/10.1109/34.709616}.

\end{thebibliography}

\clearpage
\appendix







\section{Baseline models}
\label{subsec:Baseline models}

\subsection{Classification-based Font Embedding}

Figure~\ref{fig:font-embedding-baselines} (a) shows font embedding via font classification~\cite{font-cls}.
The embedded font  $\hat{f}$ is passed to a single fully connected layer $F_{cls}$ to classify a glyph into a class in a given font set.
If the model is trained with $N_f$ fonts, then the final output is $N_f$-dimensional one-hot vector.
We used cross-entropy loss to train the model to classify glyphs $\{g|g \in \mathbb{G}_{f_k}\}$ into the font class $f_k$.
Since the classification head has no use for unseen fonts, only the font embedding network $F$ was used to embed glyphs of new fonts.

\subsection{Attribute Prediction-based Font Embedding}

Figure~\ref{fig:font-embedding-baselines} (b) shows font attribute prediction~\cite{odonovan:2014:font, Chen2019LargeScaleTF}.
A font $f_i$ in the O'Donovan dataset $\{f_i| 1 \leq i \leq 120\}$ can be expressed with attributes $a_i \in [0,1]^{37}$, as described in Section~\ref{sec:datasets}.
Therefore, this model adds a single fully connected layer $F_{attr}$ to predict attributes from font embedding ${\hat{f}}$.
The final output dimension of $F_{attr}$ is the number of attributes, $N_{attr}=37$, in case of the O'Donovan dataset.
We used binary cross-entropy loss for each attribute since it performed better than $L_2$ loss.
This method has been studied in O'Donovan~\etal~\cite{odonovan:2014:font} as attribute prediction with gradient boosted regression trees~\cite{10.1214/aos/1013203451}, and Chen~\etal~\cite{Chen2019LargeScaleTF} as tag recognition.

\subsection{Autoencoder-based Font Embedding}

Figure~\ref{fig:font-embedding-baselines} (c) shows autoencoder~\cite{autoenc1, autoenc2}.
This model simply reconstructs the input with the help of the generator network $G$.
The condensed feature vector ${\hat{f}}$ should contain a high-level abstraction for good reconstruction performance.
We used $L_1$ loss for the image reconstruction.

\subsection{Style Transfer-based Font Embedding}
\label{sec:style-transfer}

Figure~\ref{fig:font-embedding-baselines} (d) shows font style transfer  or conditional autoencoder.
Unlike {\it  Autoencoder}, this model transforms original input into a different character, preserving the font style.
For example, the generator network $G$ accepts two inputs: first,  an embedding $\hat{f}$ of glyph image $g^A$ representing character ``A''; second, the one-hot vector $c^B$ representing character ``B''.
Then the output $G(F(g^A), c^B)=g^B$ has to be a glyph image representing character ``B'' but preserving the font style of  $g^A$.
Therefore, the model $F$ must capture the font style regardless of the characters expressed in the input glyph image.
This framework has been studied in various font style transfer methods~\cite{DG-Font, zhang2018separating} with respective modifications.
We used $L_1$ loss for the image generation.

\subsection{Srivatsan~\etal~\cite{srivatsan-etal-2019-deep}}

\begin{figure}[]
  \centering
  \includegraphics[width=.5\linewidth]{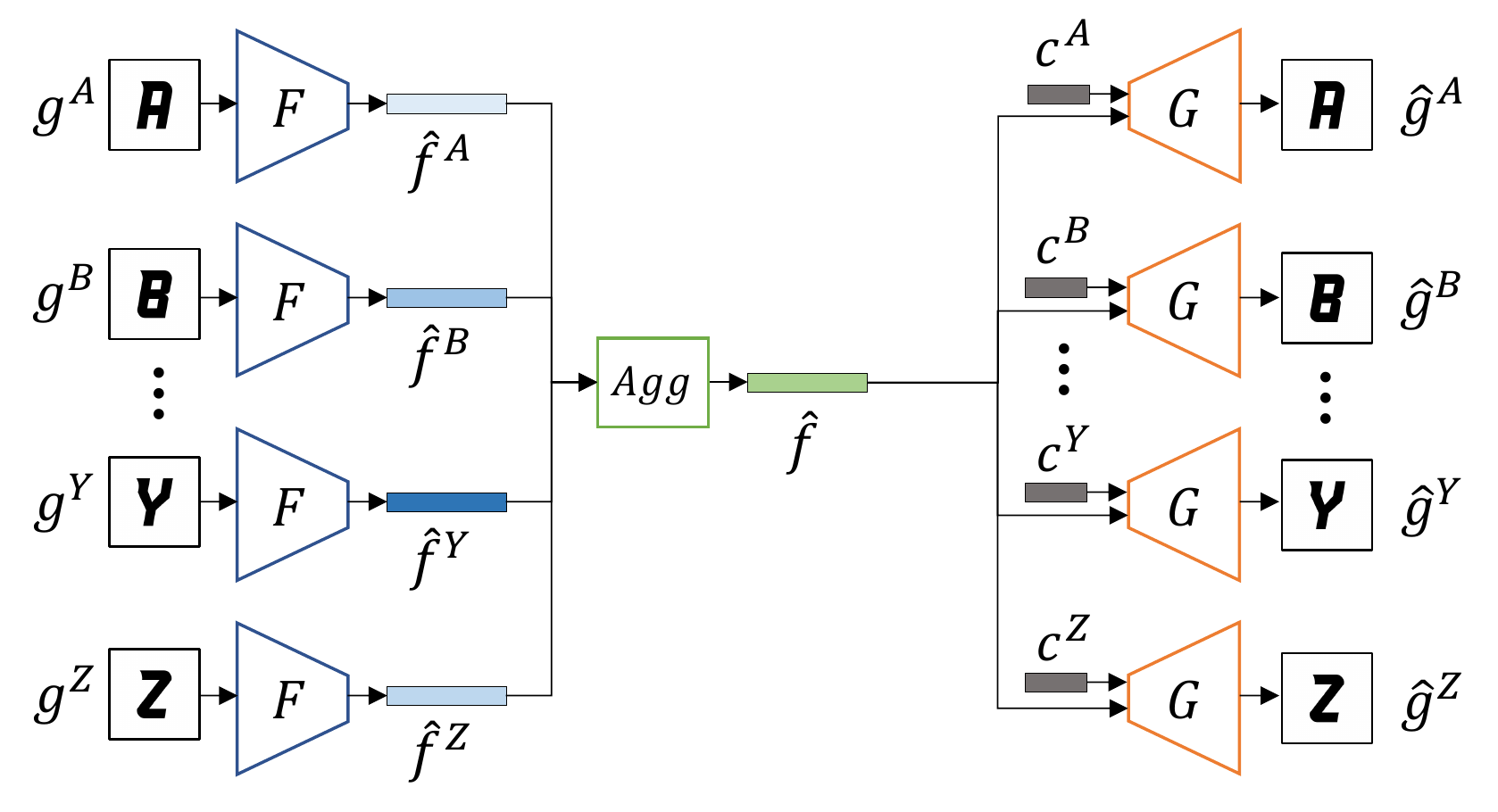}
  \caption{
    Srivatsan~\etal~\cite{srivatsan-etal-2019-deep} architecture
  }
  \label{fig:supplementary_srivatsan}
\end{figure}

Figure~\ref{fig:supplementary_srivatsan} depicts the architecture of Srivatsan~\etal~\cite{srivatsan-etal-2019-deep} method.
The model accepts glyph set ($\{ g^{\text{C}} | \text{C} \in \mathbb{C}_\text{A-Z} \}$) in a font as input, extracts features as $\{ \hat{f}^{\text{C}} | \text{C} \in \mathbb{C}_\text{A-Z} \}$ of $\texttt{feat\_dim}=1024$.
Then, aggregate them as $\hat{f}$ as follows:
\begin{equation}
    Agg(\hat{f}^{\text{C}} | \text{C} \in \mathbb{C}_\text{A-Z}) = \hat{f},
\end{equation}
where $Agg$ is a neural network.
To be more speicific, $Agg$ is $Maxpool-FC^{(1024 \times 128 )}-R-FC^{(128 \times 128)}-R-FC^{(128 \times 128)}-R-FC^{(128 \times 64)}$, and $Maxpool$ operation shrinks glyphs $26 \times 1024 $ into $1 \times 1024$.
From re-parameterizing trick~\cite{vae} on last feature dimension of $Agg$, first $32$ elements are $\mu$ and last $32$ elements are $\Sigma$ while training, and $\mu$ is considered as $\hat{f}$ when model inference.
The model then generates each glyph from $\hat{f}$ as $\{ \hat{g}^{\text{C}} | \text{C} \in \mathbb{C}_\text{A-Z} \}$.

\section{Implementation Details}
\label{sec:supp_implementation_details}
The generator $G$ architecture of {\it Style Transfer} and {\it Autoencoder} is as follows:

\begin{itemize}
    \item up0 : $TConv^{\texttt{feat\_dim}, 512}_{k=2, s=1}-IN-ReLU$,
    \item up1 : $TConv^{512, 512}_{k=4, s=2,p=1}-SA-IN-ReLU-D$,
    \item up2 : $TConv^{512, 256}_{k=4, s=2,p=1}-SA-IN-ReLU$,
    \item up3 : ${TConv}^{256, 128}_{k=4, s=2,p=1}-SA-IN-ReLU$,
    \item up4 : ${TConv}^{128, 64}_{k=4, s=2,p=1}-SA-IN-ReLU$,
    \item final : $UP_{s=2, p=1}-{Conv}^{64, 3}_{k=4, s=1,p=1}-Tanh$.
\end{itemize}
$TConv^{c_\text{in}, c_\text{out}}_{k, s, p}$ is transposed convolution, ${Conv}^{c_\text{in}, c_\text{out}}_{k, s, p}$ is convolution with $k=\mathtt{kernel\_size}$, $s=\mathtt{stride}$, $p=\mathtt{padding}$, $D$ as dropout with $p=0.5$, $UP$ as nearest neighbor upsample layer, $IN$ as instance normalization~\cite{ulyanov_instance_2016}, $SA$ as self-attention layer~\cite{Woo_2018_ECCV, zhang2018rcan}.

\newpage

\section{Evaluations of Font Embedding  Quality}
\label{subsec:Evaluation Metric}

Previous method~\cite{srivatsan-etal-2019-deep} tried to measure font embedding quality by evaluating the quality of generated font images.
We believe that an additional generator network training is not needed to evaluate font embeddings.
Moreover, accurate evaluation of a font embedding model is not possible without direct performance measurement of font representation.
Therefore, we use evaluation metrics that measure existing representation learning techniques in the font latent space via font retrieval with query glyphs.
Since the output of the encoder $F$ is a vector representation of the font latent space, we use a rank-based evaluation metric to measure the retrieval accuracy. 
Rank-based metrics have been used for in various representation learning methods in person re-identification, knowledge graph.
The similar metric, Cumulative Matching Characteristics ($\text{CMC}_k$, rank-k matching accuracy), is already popularly used in person re-identification field.

We measure font retrieval accuracy for query glyphs in the validation set $\mathbb{F}_\text{val}$.
The font retrieval task is to estimate the font of a query glyph $(g_{f_s}^{\text{C}_i})$ for a character $\text{C}_i$ by comparing it with all fonts of gallery glyph set $(\mathbb{G}_{\text{C}_j}=\{g_{f_k}^{\text{C}_j}| f_k \in \mathbb{F}_\text{val} \})$ for another character $\text{C}_j \ne \text{C}_i$. 
To find font correspondence between two glyphs, we estimate the font similarity of two glyphs based on the distance in the latent space.
Font representation vectors of the query glyph and that of a gallery glyph are extracted from the font embedding network $F$.
\begin{figure}[]
  \centering
  \includegraphics[width=0.6\linewidth]{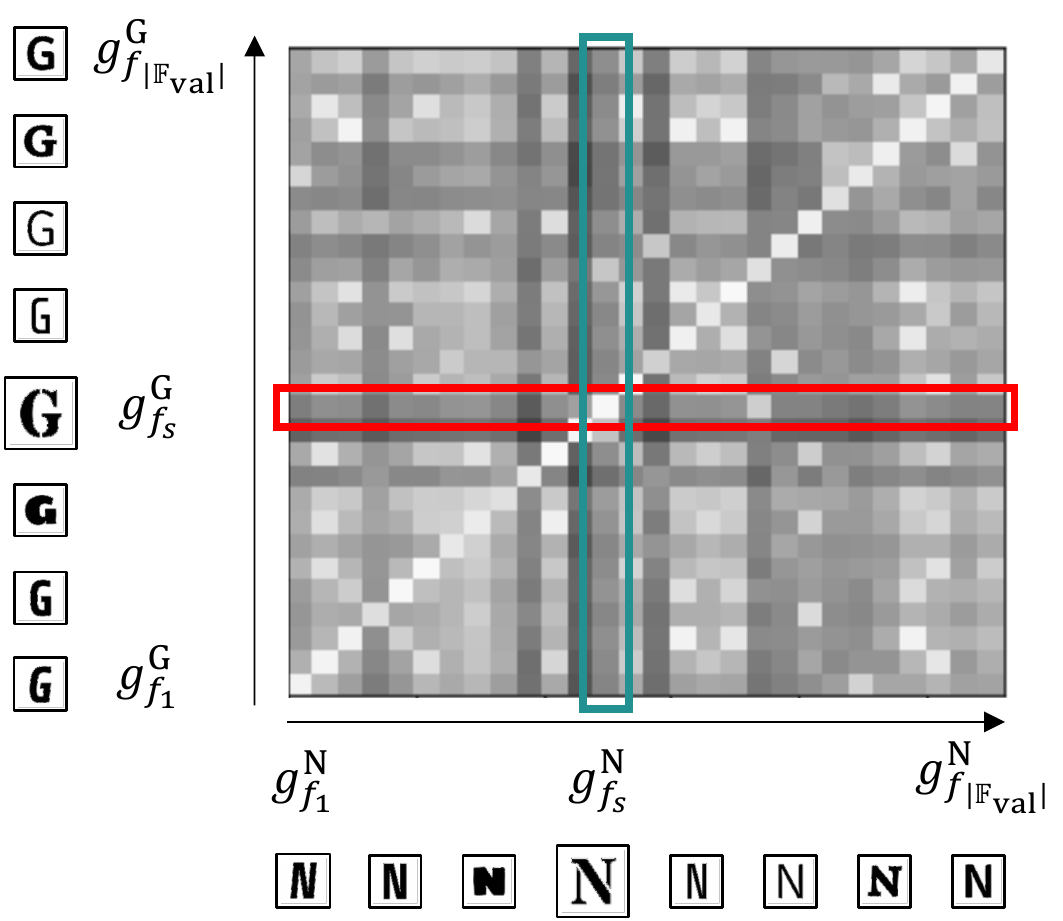}
  \caption{
  Font embedding distance matrix. 
  }
  \label{fig:character-embedding-distance}
\end{figure}

For example, for two different characters $\text{C}_i = \text{G}$ and $\text{C}_j = \text{N}$, there are two glyph sets  $\mathbb{G}_\text{G}=\{g_f^\text{G} | f \in \mathbb{F}_\text{val} \} $ and $\mathbb{G}_\text{N}=\{g_f^\text{N} | f \in \mathbb{F}_\text{val} \}$.
Figure ~\ref{fig:character-embedding-distance} shows the embedded font distance matrix between $\mathbb{G}_\text{G}$ and $\mathbb{G}_\text{N}$, of which element represents the $L_2$ embedded font distance of each glyph pair of 
$\{(g_{f_i}^\text{G}, g_{f_j}^\text{N}) | f_i, f_j \in \mathbb{F}_\text{val}\}$ in the latent space.
For a query glyph $g_{f_s}^\text{G}$ in the query set $\mathbb{G}_\text{G}$, we search a gallery glyph with the shortest font distance from the $g_{f_s}^\text{G}$, among those of all pairs $\{ (g_{f_s}^\text{G}, g)| g \in \mathbb{G}_\text{N}\}$. In Figure~\ref{fig:character-embedding-distance}, the diagonal elements show the shortest font distance in the \textcolor{red}{row-wise} sense for each query glyph in the query set $\mathbb{G}_\text{G}$.

For generality, we denote the query glyph set as  $\mathbb{G}_{\text{C}_i}=\{g_{f_k}^{\text{C}_i}| f_k \in \mathbb{F}_\text{val} \}$ and the gallery glyph set as  $\mathbb{G}_{\text{C}_j}=\{g_{f_k}^{\text{C}_i}| f_k \in \mathbb{F}_\text{val} \}$.
To evaluate the font embedding quality, we check if the embedded font $F(g_{{f}_k}^{\text{C}_i})$ of each query glyph $g_{f_k}^{\text{C}_i}$ matched with the embedded font of the gallery glyph having the same font, $F(g_{f_k}^{\text{C}_j})$, in the gallery glyph set. This implies that $ \|  F(g_{{f}_k}^{\text{C}_i}) -F(g_{{f}_k}^{\text{C}_j} )   \| $ is the lowest distance among all pairs $\{ (g_{f_k}^{\text{C}_i}, g)| g \in \mathbb{G}_{\text{C}_j}\}$ ). 
By counting these matches and dividing the total by $|\mathbb{F}_\text{val} |$,
the retrieval accuracy for a query glyph set $\mathbb{G}_{\text{C}_i}$ from a gallery glyph set $\mathbb{G}_{\text{C}_j}$ is defined by
\begin{equation}
    \underset{Ret}{\texttt{\textbf{ACC}}}
    \left(\mathbb{G}_{\text{C}_i}, \mathbb{G}_{\text{C}_j}\right)  =  
    \dfrac
    {1}{|\mathbb{F}_\text{val}|}  
    \left(\texttt{\# of query-gallery matches}\right).
\end{equation}
As the number of fonts in the validation set ($|\mathbb{F}_\text{val}|$) increases, the font retrieval becomes more challenging.
The retrieval accuracy
$ \underset{Ret}{\texttt{\textbf{ACC}}}
    (\mathbb{G}_{\text{C}_j}, \mathbb{G}_{\text{C}_i})$
and 
$ \underset{Ret}{\texttt{\textbf{ACC}}}
    (\mathbb{G}_{\text{C}_i}, \mathbb{G}_{\text{C}_j})$
are different since $\mathbb{G}_{\text{C}_i}$ and $\mathbb{G}_{\text{C}_j}$ are not identical.
For $ \underset{Ret}{\texttt{\textbf{ACC}}}
    (\mathbb{G}_{\text{C}_j}, \mathbb{G}_{\text{C}_i})$,
we consider $\mathbb{G}_{\text{C}_j}$ as the query glyph set and $\mathbb{G}_{\text{C}_i}$ as the gallery glyph set.
Thus, a diagonal element showing the shortest distance in the \textcolor{red}{row-wise} sense in Figure~\ref{fig:character-embedding-distance} may not show the shortest distance in the \textcolor{teal}{column-wise} sense.

To evaluate retrieval accuracy of all possible query set and gallery set pairs 
$ (\mathbb{G}_{\text{C}_i}, \mathbb{G}_{\text{C}_j})$ 
    in $\text{C}_i, \text{C}_j \in \mathbb{C}_\text{a-Z}$, the retrieval mean accuracy $\underset{Ret}{\texttt{\textbf{MACC}}}(\mathbb{C}_\text{a-Z})$ is defined by
\begin{equation}
    \underset{Ret}{\texttt{\textbf{MACC}}}(\mathbb{C}_\text{a-Z}) =
    \dfrac{1}{|\mathbb{C}_{\text{a-Z},2}|}\sum_{\substack{\text{C}_i, \text{C}_j \in \mathbb{C}_\text{a-Z}\\ \text{C}_i\ne \text{C}_j}} \underset{Ret}{\texttt{\textbf{ACC}}} \left(\mathbb{G}_{\text{C}_i}, \mathbb{G}_{\text{C}_j}\right),
\end{equation}
where $|\mathbb{C}_{\text{a-Z},2}| = |\mathbb{C}_\text{a-Z}| \cdot (|\mathbb{C}_\text{a-Z}|-1)$ indicates the number of pairs containing two distinct elements from the character set $\mathbb{C}_\text{a-Z}$ in an ordered manner. 

$\underset{Ret}{\texttt{\textbf{MACC}}}$ can be considered as the special case of $\text{CMC}_k$ when $k=1$ and each matching pair in query and gallery sets are glyphs of the same font but different characters.
We evaluate $\underset{Ret}{\texttt{\textbf{MACC}}}(\mathbb{C}_\text{a-Z})$ in Section~\ref{sec:Un-seen Fonts} to see how our model can well generalize to unseen fonts.

\newpage

\section{Evaluation on Unseen Characters}
\label{sec:Un-seen Characters}
\begin{table}
\centering
\caption{Performance evaluation on unseen characters.}
\label{tab:unseen-characters}
\begin{tabular}{c|c|c}
\noalign{\smallskip}\noalign{\smallskip}\hline\hline
& \multicolumn{1}{c|}{OFL valset 
} 
& \multicolumn{1}{c}{O'Donovan valset 
} \\
      &  $\underset{Ret}{\texttt{\textbf{MACC}}}(\mathbb{C}_\text{0-9} , \mathbb{C}_\text{a-Z}) $
      &  $\underset{Ret}{\texttt{\textbf{MACC}}}(\mathbb{C}_\text{0-9} , \mathbb{C}_\text{a-Z} )$ 
      \\      
\hline

\rule{0pt}{11pt}
$\hat{f}$ of  Paired-$g$ Matching  & {\bf 46.01} &   {\bf 63.21}\\
$\hat{f}$ of Classification & 42.94 &   58.74  \\
$\hat{f}$ of  Style Transfer & 25.41 &   40.85 \\
$\hat{f}$ of  Autoencoder & 12.80 &   26.76 \\

\hline
\hline
\end{tabular}

\end{table}

We can also evaluate the retrieval mean accuracy for two different character sets.
For example, in case of number query glyph set $\mathbb{G}_{\text{C}_i}$ ($\text{C}_i \in \mathbb{C}_\text{0-9}$) and alphabet gallery glyph set $\mathbb{G}_{\text{C}_j}$ ($\text{C}_j \in \mathbb{C}_\text{a-Z}$),   
the retrieval mean accuracy is defined as follows:
\begin{equation}
\label{eq:0-9->a-Z}
    \underset{Ret}{\texttt{\textbf{MACC}}}(\mathbb{C}_\text{0-9}, \mathbb{C}_\text{a-Z})= 
    \dfrac{1}{|\mathbb{C}_\text{0-9}| \cdot |\mathbb{C}_\text{a-Z}|}\sum_{\substack{\text{C}_i \in \mathbb{C}_\text{0-9}\\ \text{C}_j \in \mathbb{C}_\text{a-Z}}}
    \underset{Ret}{\texttt{\textbf{Acc}}} \left(\mathbb{G}_{\text{C}_i}, \mathbb{G}_{\text{C}_j}\right).
\end{equation}

We use the retrieval mean accuracy for two different character sets to see how well our model generalizes to glyphs of unseen characters.
Glyphs representing $\mathbb{C}_\text{0-9}$ were used for evaluation to measure font embedding generalization to unseen characters.
Our models were trained with alphabet set $\mathbb{C}_\text{a-Z}$ and had not seen number set $\mathbb{C}_\text{0-9}$.
Using characters in $\mathbb{C}_\text{0-9}$ as the query set and characters in $\mathbb{C}_\text{a-Z}$ as the gallery set, we present each font embedding method evaluated with the retrieval mean accuracy $\underset{Ret}{\texttt{\textbf{MACC}}}(\mathbb{C}_\text{0-9}, \mathbb{C}_\text{a-Z})$~(Eq~\ref{eq:0-9->a-Z}) in Table~\ref{tab:unseen-characters}.
As discussed in Section~\ref{sec:Un-seen Fonts}, we found the best model performed on the O'Donovan validation set and reported on the OFL dataset and vice versa.
Our {\it Paired-glyph Matching} again achieved the best score compared to other font embedding methods.

\newpage

\section{More O'Donovan dataset embeddings}

\begin{figure*}[]\centering
\includegraphics[width=1\linewidth]{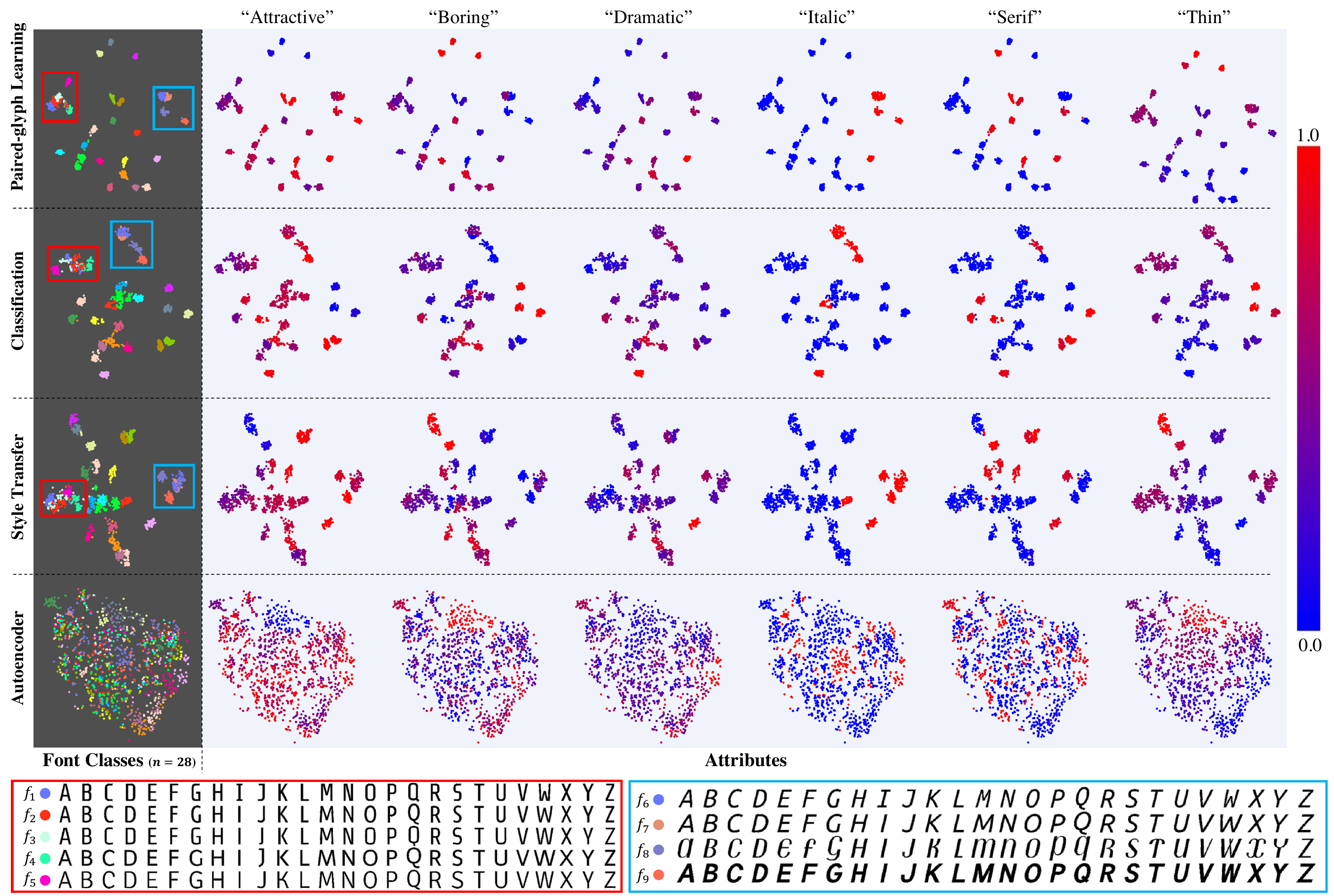}
\vspace*{-0.3cm}
\caption{Font latent space on the O'Donovan dataset annotated by font  attributes.
} 
\label{fig:supplementary_donovan_embeddings_attributes}
\vspace{-0.3cm}
\end{figure*}

Figure~\ref{fig:supplementary_donovan_embeddings_attributes} extends Figure~\ref{fig:donovan-font-embeddings} and shows more $\hat{f}$ visualization on the O'Donovan validation set with attributes.



\end{document}